\documentclass[11pt,a4paper]{article}
\usepackage[hyperref]{acl2020}
\usepackage{times}
\usepackage{latexsym}

\usepackage{microtype}

\usepackage{array, caption, floatrow, tabularx, makecell, booktabs}%
\usepackage{hyperref}
\usepackage{color}
\usepackage{colortbl}
\usepackage[colorlinks]{}
\usepackage{multirow}
\usepackage{graphicx}
\usepackage{caption}
\usepackage{subcaption}
\usepackage{url}
\usepackage{fixltx2e}
\usepackage{scalefnt}
\usepackage{xcolor}
\usepackage{dblfloatfix}
\definecolor{darkred}{RGB}{139,0,0}

\usepackage{algorithm2e}

\usepackage{pgfplots}

\makeatletter
\def\BState{\State\hskip-\ALG@thistlm}
\makeatother

\usepackage{url}
\aclfinalcopy 

\title{Learning from Mistakes: \\ Combining Ontologies via Self-Training for Dialogue Generation}

\author{Lena Reed$^1$, Vrindavan Harrison$^1$, Shereen Oraby$^2$\thanks{$\:\:$Work done prior to joining Amazon.},  \\ 
\bf Dilek Hakkani-T{\"u}r$^{2*}$, \bf and Marilyn Walker$^1$ \\
$^1$Natural Language and Dialogue Systems Lab, University of California, Santa Cruz\\
$^2$Amazon Alexa AI\\
  {\tt \{lireed,vharriso,mawalker\}@ucsc.edu} \\
  {\tt \{orabys,hakkanit\}@amazon.com} \\}

\date{}

\begin{document}
\maketitle

\begin{abstract}
Natural language generators (NLGs) for task-oriented dialogue typically
take a meaning representation (MR) as input,
and are trained end-to-end with a corpus of MR/utterance
pairs, where the MRs cover a specific set of dialogue acts and domain
attributes. Creation of such datasets is labor intensive and time
consuming. Therefore, dialogue systems for new domain ontologies would benefit from using data for pre-existing ontologies.  
Here we explore, for the first time, whether  it  is possible
to train  an NLG for a new {\bf larger}
ontology using  existing training sets for the restaurant domain, where each set is based on a {\bf different} ontology. We create a new, larger {\bf combined}  ontology,  and then  train an NLG  to produce utterances covering it.
For example, if one dataset has  attributes for {\it family friendly} and {\it
  rating} information, and the other has attributes for {\it decor}
and {\it service}, our aim is an NLG for the combined ontology
that can produce utterances that
realize values for {\it family friendly}, {\it rating}, {\it decor} and {\it
  service}.  
Initial experiments with a baseline neural sequence-to-sequence model
show that this task is surprisingly  challenging.
We then develop a  novel {\bf self-training} method that
identifies (errorful) model outputs, automatically 
constructs a corrected MR input to form a new (MR, utterance) training pair, and then repeatedly adds these new instances back into the training data.
We then test the resulting model on a new test set. The result is a self-trained
model whose performance is an absolute 75.4\% improvement over the baseline model. 
We also report a human qualitative evaluation of the final  model showing that it achieves high
naturalness, semantic coherence and grammaticality. 
\end{abstract}

\section{Introduction}
Natural language generators (NLGs) for task-oriented dialogue take
meaning representations (MRs) as inputs, i.e. a set of dialogue acts
with attributes and their values, and output natural language
utterances realizing the MR.  Current NLGs are trained end-to-end with
a corpus of MR/utterance pairs where the MRs cover a specific set of
dialogue acts and domain attributes. Creation of such datasets is
labor intensive and time consuming. However, when building an NLG for
a new domain ontology, it should be possible to re-use data built on
existing domain ontologies.  If this were possible, it would speed up
development of new dialogue systems significantly.

\begin{figure*}[t!bh]
\begin{footnotesize}
\begin{tabular}{p{.15in}|p{.49in}|p{2.8in}|p{2.2in}} \toprule
\bf ID & \bf Ontology & \bf MEANING REPRESENTATION & \bf EXAMPLE \\ \hline 
\sc e1 & \sc nyc (training)  & \sc {\color{darkred}recommend[yes]}, inform({\color{darkred}name[restaurant]}, {\color{darkred}service[excellent], food[excellent], d\'ecor[excellent]}, location[area], {\color{darkred}price[expensive]}) &
\underline {{\color{darkred}I suggest you go to [{\sc restaurant}]}}. The {\color{darkred}\underline {food, service} and \underline{atmosphere} \underline{are all excellent}}, even if it is {\color{darkred}expensive}. Its in [{\sc area}].   \\ \hline

\sc e2 &  \sc e2e (training) & 
\sc  inform(name[restaurant], {\color{blue}eatType[restaurant-type], 
  customer-rating[high]}, area[area],
   {\color{blue}near[point-of-interest]})  & 
[{\sc restaurant}] is a {\color{blue}\underline{[{\sc restaurant-type}]}} in [{\sc area}] {\color{blue}\underline{near [{\sc point-of-interest}]}}. It has a {\color{blue}\underline{high customer rating}}.   \\ \hline

\sc e3 & \sc combined (test) & 
\sc {\color{darkred}recommend = yes}, inform({\color{darkred}name[restaurant]}, {\color{blue}eatType[restaurant-type]}, 
{\color{blue}food = excellent}, location[area], {\color{blue}near[point-of-interest]}, {\color{blue}customer-rating[high]}, {\color{darkred}d\'ecor = excellent, service=excellent, price=expensive}) & 
{\color{darkred}[{\sc restaurant}] is the best} because it has {\color{darkred}excellent service and atmosphere}. 
It is a {\color{blue}[{\sc restaurant-type}]} offering {\color{darkred}excellent food} in [{\sc area}] {\color{blue}near [{\sc point-of-interest}]} with a {\color{blue}high customer rating}, but it is {\color{darkred}expensive}. \\ 
\bottomrule
\end{tabular}
\end{footnotesize}
\vspace{-.1in}
 \caption{\label{blend-example} E1 and E2 illustrate  training
  instances from the two source datasets E2E and NYC. E2E attributes are represented in {\color{blue}blue} and NYC is in {\color{darkred}red}. Some attributes are shared
   between both sources: here the unique dialogue acts and attributes
   for each source are underlined in E1 and E2.  E3 illustrates an MR
   from the target test set that we dub COM. All the MRs in COM combine dialogue acts
   and attributes from E2E and NYC. There is no training data
    corresponding to E3.
   The MRs
   illustrate how some attribute values, e.g. {\sc restaurant name,
     point-of-interest}, are delexicalized to improve generalization.}

\end{figure*}

Here we experiment with one version of this task by building a new
domain ontology based on {\bf combining} two existing ontologies, and
utilizing their training data.  Each dataset is based on a different domain ontology in the restaurant domain, with novel attributes and dialogue acts not seen in the other dataset, e.g. only one has attributes representing {\it family friendly} and {\it rating} information, and only one has attributes for {\it decor} and {\it service}.  Our aim is an NLG engine that can realize utterances for the extended {\bf
  combined} ontology not seen in the training data, e.g. for MRs that
specify values for {\it family friendly}, {\it rating}, {\it decor} and {\it
  service}.  Figure~\ref{blend-example} illustrates this task.
Example E1 is from a training set referred to as NYC, from previous
work on controllable sentence planning in NLG \cite{reed2018can},
while E2 is from the E2E NLG shared task \cite{Novikovaetal17}. As we
describe in detail in Section~\ref{ontology-sec}, E1 and E2 are
based on two distinct ontologies.  Example E3 
illustrates the task addressed in this
paper: we create a test set of novel MRs for the combined ontology,
and train a model to generate high quality outputs where individual
sentences realize attributes from both ontologies.

To our knowledge, this is a completely novel task.  While it is common
practice in NLG to construct test sets of MRs that realize attribute
combinations not seen in training, initial experiments showed that
this task is surprisingly adversarial.  However, methods for supporting this
type of generalization and extension to new cases would be of great
benefit to  task-oriented dialogue systems, where it is 
common to start with a restricted
set of attributes and then enlarge the domain ontology over time. New
attributes are constantly being added to databases of restaurants,
hotels and other entities to support better recommendations and better
search.  Our experiments test whether existing data that only covers a
subset of attributes can be used to produce an NLG for the enlarged
ontology.

We describe below how we create a test set --- that we call {\sc com}
--- of combined MRs to test different methods for creating such an
NLG.  A baseline sequence-to-sequence NLG model
has a slot error rate (SER) of .45
and only produces semantically perfect outputs 3.5\% of the time. To
improve performance, we experiment with three different ways of
conditioning the model by incorporating {\it side constraints} that
encode the source of the attributes in the MR
\cite{Sennrich_Haddow_Birch_2016,harrison2019maximizing}.
However, this only
  increases the proportion of semantically perfect model outputs from
  3.5\% to 5.5\% (Section~\ref{model-results-sec}).

We then propose and motivate a novel self-training method that greatly
improves performance by learning from the model mistakes. An error analysis
shows that the models {\bf do} produce many {\bf combined} outputs,
but with errorful semantics. We develop a rule-based text-to-meaning semantic
extractor that automatically creates novel correct MR/text training
instances from errorful model outputs, and use these in self-training
experiments, thus learning from our mistakes
(Section~\ref{self-train-sec}). We validate the text-to-meaning extractor with a human evaluation.  We find that a model trained with
this process produces SERs of only .03, and semantically perfect
outputs 81\% of the time (a 75.4 percent improvement).  A human
evaluation shows that these outputs are also natural, coherent and
grammatical. Our contributions are:
\begin{itemize}
\item Definition  of a novel generalization task
for neural NLG engines, that of generating from unseen MRs that combine
attributes from two  datasets with different ontologies;
\vspace{-.1in}
\item Systematic experiments on methods for conditioning
NLG models, with results
showing the effects on model performance for both semantic errors
and combining attributes; 
\vspace{-.1in}
\item A novel self-training method that learns from the model's mistakes
to produce semantically correct
outputs 81\% of the time, an absolute 75.4\% improvement.
\end{itemize}

We start in Section~\ref{ontology-sec} by defining the task in more
detail, describe our models and metrics in
Section~\ref{exp-overview-sec}, and results in
Section~\ref{results-sec}.  We discuss related work throughout the
paper where it is most relevant and in the
conclusion in Section~\ref{conc-sec}.

\section{Ontology Merging and Data Curation}
\label{ontology-sec}

We start with two existing datasets, NYC and E2E, representing
different ontologies for the restaurant domain.  The NYC dataset
consists of 38K utterances \cite{reed2018can,Orabyetal18}, based on a
restaurant ontology used by Zagat
\cite{Stentetal02,StentPrasadWalker04}.\footnote{{{http://nlds.soe.ucsc.edu/sentence-planning-NLG}}}
The E2E dataset consists of 47K utterances distributed for the E2E
Generation Challenge
\cite{Novikovaetal17}.\footnote{{{http://www.macs.hw.ac.uk/InteractionLab/E2E/}}}
Each dataset consists of pairs of reference utterances and meaning
representations (MRs).  Figure~\ref{blend-example} shows sample MRs
for each source and corresponding training instances as E1 and E2.

\begin{figure}[h!tb]
\centering
\includegraphics[width=2.35in]{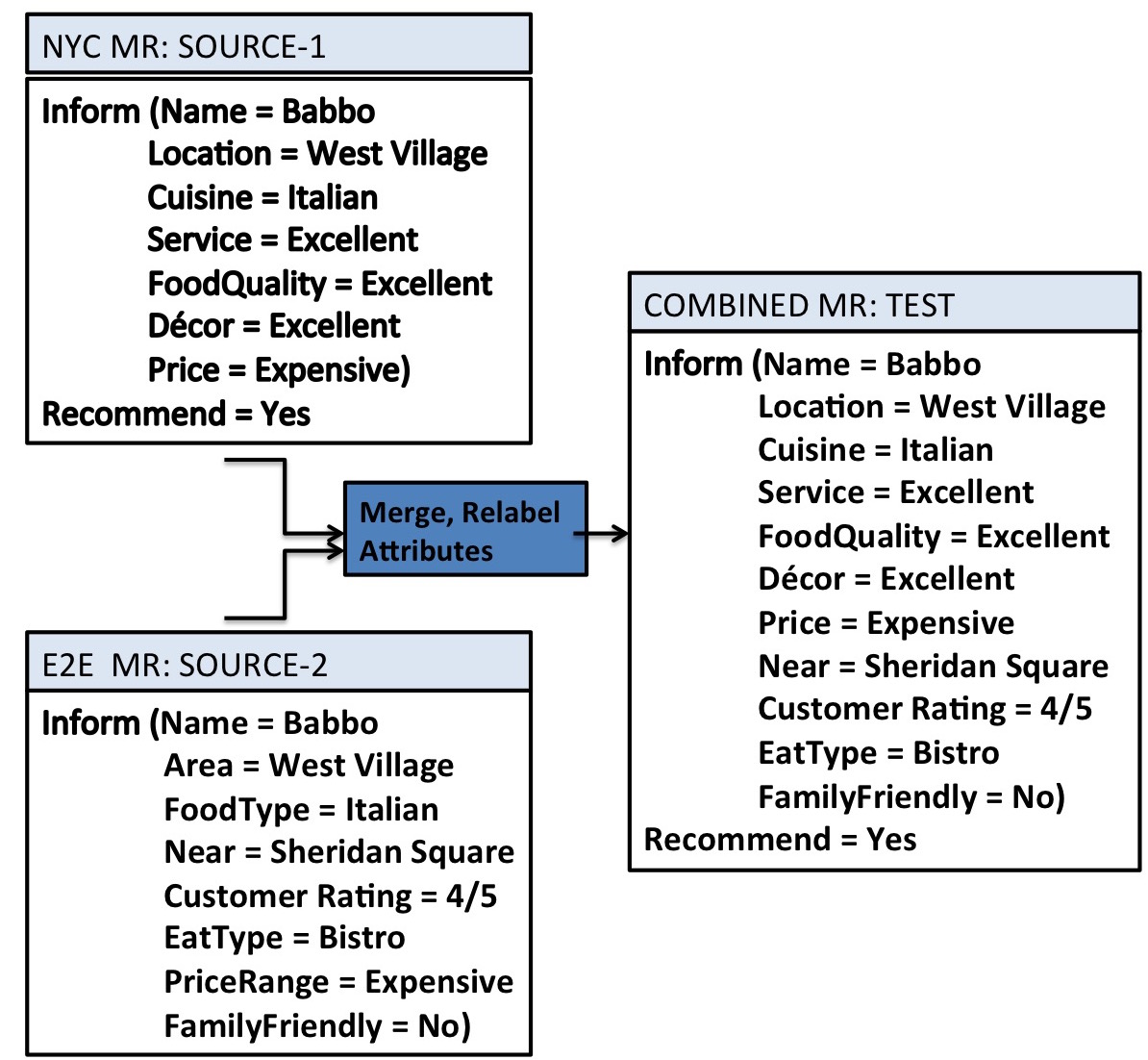}
\vspace{-.1in}
\caption{\label{ontology-fig} An example illustrating
how dialogue acts and attributes for both source databases are merged and relabelled
to make a new combined ontology used in train and test.}
\end{figure}

\noindent{\bf Ontology Merging.} We first   make a new combined ontology {\sc onto-com} by merging NYC and E2E. Attributes, dialogue acts, and
sample values for E2E and NYC are illustrated on the left-hand side of
Figure~\ref{ontology-fig}, and the result of merging them to create
the new ontology is on the right-hand side of
Figure~\ref{ontology-fig}. Since there are only 8 attributes in each
source dataset, we developed a script by hand that maps the MRs from
each source into the {\sc onto-com} ontology.

As Figure~\ref{ontology-fig} shows, both datasets have the {\sc
  inform} dialogue act, and include the attributes {\it name}, {\it
  cuisine}, {\it location}, and {\it price} after mapping.  The unique
attributes for the NYC ontology are scalar ratings for {\it service},
{\it food quality} and {\it decor}.  The NYC dataset also has the {\sc
  recommend} dialogue act, seen in E1 in
Figure~\ref{blend-example}. The unique attributes of the E2E ontology
are {\it customer rating}, {\it eat type} (``coffee shop''), {\it near} and {\it family friendly}.

\noindent{\bf Training Data.} Given the combined ontology {\sc onto-com},
we then map the training data for both E2E and NYC into {\sc onto-com}
by relabelling the MRs to have
consistent names for shared attributes as illustrated in
Figure~\ref{ontology-fig}. We create a balanced training set of
$\sim$77K from the two original datasets by combining all NYC references
with a random same-size sample of E2E references.

\noindent{\bf Test Set.} We then manually
create a test set, COM, consisting of 3040 MRs based on the
new combined ontology {\sc onto-com}. Each test MR must have at least one attribute from E2E and one attribute from NYC so that it combines 
attributes from both sources: these MRs
provide combinations never seen in
training.\footnote{The train and test data are available at http://nlds.soe.ucsc.edu/source-blending-NLG}  Example E3 in Figure~\ref{blend-example} provides an example 
test MR.
The procedure for creating the test set ensures that the  length and complexity  of the test set are
systematically varied, with lengths normally distributed and ranging from 3 to 10 attributes. Recommendations only occur in the NYC training data, and they increase both {\bf semantic} and {\bf syntactic} complexity, with longer utterances that use the discourse relation of {\sc justification} \cite{Stentetal02}, e.g. {\it Babbo is the best {\bf because} it
  has excellent food}. We hypothesize that recommendations  may be more
challenging to combine across domains,  so we vary MR complexity by
including the {\sc recommend} dialogue act in half  the test
references. We show in Section~\ref{results-sec} that the length
and complexity of the MRs is an important factor in the performance
of the trained models.

\section{Experimental Overview and Methods}
\label{exp-overview-sec}

Given the training and test sets for the combined ontology in
Section~\ref{ontology-sec}, we test 4 different neural model
architectures and present results in
Section~\ref{model-results-sec}. We then propose a a novel
self-training method, and present results in
Section~\ref{self-train-sec}.  These experiments rely on the model
architectures presented here in Section~\ref{models-sec}, and the Text-to-Meaning semantic extractor and
performance metrics 
in Section~\ref{ttm-sec}.

\subsection{Model Architectures}
\label{models-sec}

In the recent E2E NLG Challenge shared task, models were tasked with
generating surface forms from structured meaning
representations (MRs) \cite{duvsek2020evaluating}. The top performing
models were all RNN encoder-decoder systems.
Here we also use  a standard RNN Encoder--Decoder model
\cite{sutskever2014sequence} that maps a source sequence (the input
MR) to a target sequence (the utterance text).  We first implement a baseline
model and then add three variations of model supervision that aim to
improve semantic accuracy. 
All of the models are built with
OpenNMT-py, 
a sequence-to-sequence modeling framework \cite{klein2017opennmt}.

\noindent
\textbf{Encoder.} 
The MR is represented as a sequence of (attribute, value) pairs with separate vocabularies for attributes and values. Each attribute and each value are represented using 1-hot 
vectors. An (attribute, value) pair is represented by concatenating the two 1-hot vectors.

The input sequence is processed using two single layer
bidirectional-LSTM \cite{hochreiter1997long} encoders.  The first
encoder operates at the pair level, producing a hidden state for each
attribute-value pair of the input sequence. The second LSTM encoder is
intended to produce utterance level context information in the form of
a full MR encoding produced by taking the final hidden state after
processing the full input sequence.  The outputs of both encoders are
combined via concatenation. That is, the final state of the second
encoder is concatenated onto each hidden state output by the first
encoder.  The size of the pair level encoder is 46 units and the size
of the MR encoder is 20 units.  Model parameters are initialized using
Glorot initialization \cite{glorot2010understanding} and optimized
using Stochastic Gradient Descent with mini-batches of size 128.

\noindent \textbf{Decoder.} The decoder is a uni-directional LSTM that
uses global attention with input-feeding. Attention weights are
calculated via the \textit{general} scoring method
\cite{luong2015effective}.  The decoder takes two inputs at each time
step: the word embedding of the previous time step, and the attention
weighted average of the encoder hidden states. The ground-truth
previous word is used when training, and the predicted previous word when evaluating. Beam search with five beams is used during
inference.

\begin{figure}[]
	\centering
    \includegraphics[width=\columnwidth, keepaspectratio]{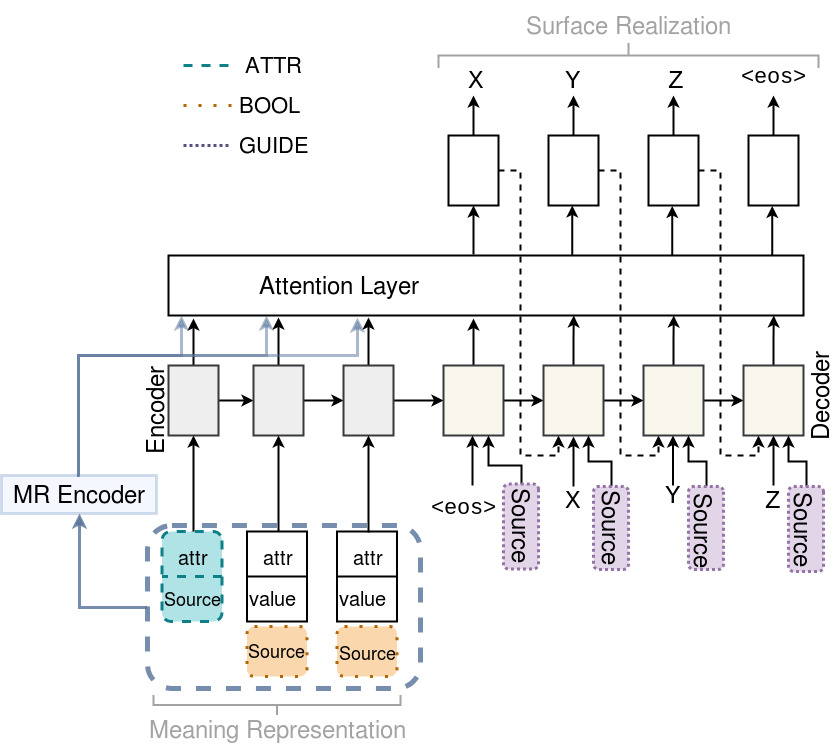}
    \caption{Attentional Encoder-Decoder architecture with each supervision method shown. 
    \label{fig:nn-side-constraints}}
    \vspace{-.1in}
\end{figure}

\begin{figure}[]
	\centering
    \includegraphics[width=0.78\columnwidth, keepaspectratio]{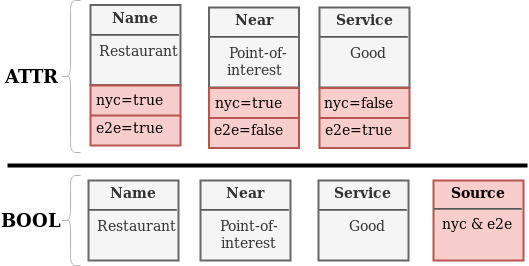}
    \caption{An illustration of {\sc attr} and {\sc bool} supervision methods, 
    with the source supervision (NYC or E2E) shown in red. 
    \label{fig:mr-attr-bool-example}}
    \vspace{-.1in}
\end{figure}

\noindent \textbf{Supervision.} Figure~\ref{fig:nn-side-constraints}
shows the baseline system architecture as well as three
types of supervision, based on conditioning on source (E2E, NYC) information. The additional supervision is
intended to help the model attend to the source
domain information.
We call the three types of supervision  {\sc guide},
{\sc attr} and {\sc bool}, and the baseline
architecture {\sc nosup}, representing that it has no additional
supervision.  

The supervision methods are shown in Figure \ref{fig:mr-attr-bool-example}.
The source feature
has a vocabulary of three items: {\it nyc}, {\it e2e} and {\it both}.
Since {\it both} is never seen in train, the source information
is represented using two booleans: {\it True$||$False} denotes a reference from E2E  while {\it False$||$True} denotes a reference from NYC. This encoding is intended to encourage generalization at inference time.  During
inference, blending of information from both sources is specified by
using {\it True$||$True}. The {\sc attr} supervision method represents the source information by
concatenating the boolean source token onto each attribute as seen in Figure~\ref{fig:mr-attr-bool-example}.
 This redundantly represents the source information locally to each attribute, which has been effective for tasks such as question generation and stylistic control \cite{harrison2018neural,harrison2019maximizing}.  The {\sc bool} supervision method adds the boolean source token to the end of the sequence of attribute-value pairs as its own attribute,  as in work on machine translation and controllable stylistic generation \cite{Sennrich_Haddow_Birch_2016,yamagishi2016controlling,Ficler_Goldberg_2017}. The {\sc guide} model inputs
the  source information directly to the decoder LSTM.  
In previous work, putting information into the decoder in this way has yielded improvements   in  paraphrase generation and  controllable generation \cite{Iyyeretal18,harrison2019maximizing}

\subsection{Text-to-Meaning Semantic Extractor}
\label{ttm-sec}

Much previous work in NLG relies on a test set that provides
gold reference outputs, and then applies automatic metrics such as
BLEU that compare the gold reference to the model output
\cite{Papinenietal02,duvsek2020evaluating}, even though  the 
limitations of BLEU for NLG are widely acknowledged
\cite{BelzReiter06,stent2005evaluating,NovikovaetalEVAL17,Liuetal16b}. To address these limitations, recent
work has started to develop 
``referenceless'' NLG evaluation metrics
\cite{dusek2018referenceless,kann2018sentence,tian2018treat,MehriEskenazi-FED}.
  
Since there are no reference outputs for the COM test set, we need a referenceless evaluation metric.
We develop a rule-based
text-to-MR semantic extractor (TTM) that allows
us to compare  the input MR  to
an  MR automatically constructed from an NLG model textual output by the TTM, in order to 
calculate {\bf SER}, the slot error rate.  The TTM system is based on information extraction methods. We conduct a human evaluation of  its accuracy below.
A similar approach is used to calculate semantic accuracy in 
other work in NLG, including comparative system evaluation in the E2E
Generation Challenge
\cite{Juraskaetal18,duvsek2020evaluating,WisemanSR17,shen2019pragmatically}.


The TTM relies on a rule-based automatic
aligner that tags each output utterance with the attributes and values
that it realizes. The aligner
takes advantage of the fact that the {\sc recommend} dialogue act, and
the attributes and their values are typically realized from a domain-specific finite
vocabulary.  The output of the aligner is then used by the TTM
extractor to construct an MR that matches the (potentially errorful)
utterance that was generated by the NLG.  We refer to this MR as the
``retrofit MR". 
 The retrofit MR is then compared to the input MR
in order to automatically calculate the slot error rate  {\bf SER}:
\begin{center}
${\displaystyle {\mathit {SER}}={\frac {D+R+S+H}{N}}}$
\end{center}
\vspace{.1in}
where $D$ is the number of deletions, $R$ is the number of 
repetitions, $S$ is the number of substitutions, $H$ is the number of
hallucinations and $N$ is the number of slots in the input MR
\cite{Nayaketal17,reed2018can,wenetal15}.
  Section~\ref{semantic-error-supplement} in the supplementary
  materials provides more detail and examples for each type of
  semantic error.
     SER is first calculated
on individual utterances and then averaged over the whole test set.
For
additional insight, we also report the percentage of {\bf semantically
perfect outputs} (perfect\%),  outputs where the SER is 0 and there are no semantic errors. This measure is analogous to the Sentence Error Rate used in speech recognition.

\noindent{\bf Human TTM Accuracy Evaluation.} We evaluated the
TTM and the automatic SER calculation with a separate
experiment where two NLG experts hand-labelled a random sample of 200
model outputs.  Over the 200 samples, the automatic SER was .45 and
the human was .46. The overall correlation of the automatic SER with
the human SER over all types of errors (D,R,S,H) is .80 and the
correlation with deletions, the most frequent error type, is .97.

 \noindent{\bf Retrofit MRs for Self-Training.} The TTM  is critical for our novel self-training method  described in Section~\ref{self-train-sec}. The retrofit MRs  match the (errorful) NLG output: when these MR/NLG output pairs combine attributes from both sources, they provide novel corrected examples
to add back into training.

\section{Results}
\label{results-sec}

We run two sets of experiments. We first run all of the NLG models
described in Section~\ref{models-sec} on the COM test set, and
automatically calculate SER and perfect\% as described in
Section~\ref{ttm-sec}.  We report these results in
Section~\ref{model-results-sec}. Section~\ref{self-train-sec} motivates and  describes the self-training method and 
presents the results, resulting in  final
models that generate semantically perfect outputs 83\% of the
time.

\subsection{Initial Model Results} 
\label{model-results-sec}


\begin{table}[!htb]
\begin{footnotesize}
\begin{tabular}
{@{} l|l|l||r|cc @{}}
\hline
 \bf Model   & \bf Training & \bf Test & {\cellcolor[gray]{0.9} {\sc ser}} & 
\multicolumn{2}{c}{\cellcolor[gray]{0.9} {\sc perfect}} \\
& & & { \cellcolor[gray]{0.9}}  & { \cellcolor[gray]{0.9} N} & { \cellcolor[gray]{0.9} \%}
\\ \hline  
\sc nosup &  \sc e2e + nyc & \sc com & \bf .45 & 106 & 3.5\% \\ 
\sc guide &   \sc e2e + nyc & \sc com & .66 & 15 & 0.5\%   \\ 
\sc attr &   \sc e2e + nyc & \sc com & .46 & 167 & \bf 5.5\%   \\ 
\sc bool &   \sc e2e + nyc & \sc com & \bf .45 & 86 & 2.8\%    \\ \hline
\end{tabular}
\end{footnotesize}
\vspace{-.1in}
\centering 
\caption{SER and perfect\% on test for each model type on the test of
3040 MRs ({\sc com}) that combine attributes from both sources.}
\label{table:supervision-evaluation}
\end{table}
\vspace{-.1in}

\noindent{\bf Semantic Accuracy.}  Table \ref{table:supervision-evaluation} summarizes the results 
across the four models {\sc nosup}, {\sc
  guide}, {\sc attr} and {\sc bool}.  
Overall, the results show that the task, and the COM test set,
are surprisingly adversarial. 
All of the models have extremely high SER, and the SER for {\sc nosup},
{\sc attr}, and {\sc bool} are very similar.  Row 2 shows that the {\sc
  guide} model has much worse performance than the other models, in
contrast to other tasks 
\cite{Iyyeretal18}. We do not examine the {\sc guide} model further. Row 3 shows that the {\sc
  attr} supervision results in the largest percentage of perfect
outputs (5.5\%).  

\begin{table}[!htb]
\begin{footnotesize}
\begin{tabular}
{@{} l|l|l||r|r @{}}
\hline
\bf Model & \bf Training  & \bf Test   &{ \cellcolor[gray]{0.9} {\sc ser}} & { \cellcolor[gray]{0.9} {\sc perf} \%}
\\ \hline  
\sc nosup & \sc e2e & \sc e2e & .16 & 19\%  \\ 
\sc nosup & \sc e2e + nyc & \sc e2e &  .18 & 15\%    \\ 
\sc nosup & \sc nyc & \sc  nyc  & .06   & 69\% \\ 
\sc nosup & \sc e2e + nyc & \sc nyc & .06 & 71\%     \\ \hline
\end{tabular}
\end{footnotesize}
\vspace{-.1in}
\centering 
\caption{Baseline results for each source on its own test using the {\sc nosup} model.
E2E test N = 630. NYC test N = 314.}
\label{table:baselines-combo-nocombo}
\end{table}

The results in Table~\ref{table:supervision-evaluation} should be compared
with the baselines
for testing {\sc nosup} on {\bf only} E2E or NYC  in Table~\ref{table:baselines-combo-nocombo}. 
Both the E2E and NYC test
sets consist of unseen inputs, where E2E is the standard E2E
generation challenge test \cite{duvsek2020evaluating}, and NYC consists of 
novel MRs with baseline attribute frequencies
matching the training data.\footnote{Previous work on the E2E dataset has also used seq2seq
  models, with SOA results for SER of 1\% \cite{duvsek2020evaluating}, but here we do not use the full training set. 
  Our partition of the NYC dataset has not been used before, but
  experiments on comparable NYC datasets have SERs of .06 and
  .02 \cite{reed2018can,harrison2019maximizing}.}  Rows 1 and 3 test models trained on only
E2E or only NYC, while Rows 2 and 4 test the same trained {\sc nosup}
model used in Row 1 of Table \ref{table:supervision-evaluation} on
 E2E or NYC test sets respectively. 
Comparing Rows 1 and 2 shows that training on the same combined data used in Table \ref{table:supervision-evaluation} slightly degrades
performance on  E2E, however, this
SER is still considerably lower than the .45 SER for the 
{\sc nosup} model tested on the COM test set, 
shown in the first row of Table
\ref{table:supervision-evaluation}. Row 4 shows that the
{\sc nosup} model trained on the combined data appears to improve 
performance on the NYC test because the perfect\% goes up from 69\% in Row 3 to
71\%.  The SER of .06 shown in Row 4 
should also be compared to the .45 SER reported
for the {\sc nosup} model in the first row of Table
\ref{table:supervision-evaluation}. 
These results taken together
establish that the combined MRs in the COM test provide a very different challenge than the E2E and NYC unseen test inputs.

However, despite the poor performance of the
initial models, we hypothesized that there may be enough good outputs to
experiment with self-training.  Since the original training data had no combined
outputs, decoding may benefit from even small numbers of training items added back in
self-training.
\begin{table}[!bht]
\begin{footnotesize}
\begin{tabular}
{@{} l||ccc @{}}
\hline
 \bf Model   &{ \cellcolor[gray]{0.9} {\sc Nat.}} & { \cellcolor[gray]{0.9} {\sc Coher.}} & { \cellcolor[gray]{0.9}  {\sc Grammat. } }
\\ \hline  
\sc nosup & 4.04 &  4.13 & 4.12 \\ 
\sc attr & \bf 4.11  &  \bf 4.25 &   4.14  \\ 
\sc bool & 3.97 & 4.18 & \bf 4.25    \\ \hline
\sc agreement & .63 & .62 & .65 \\ 

\hline
\end{tabular}
\end{footnotesize}
\vspace{-.1in}
\centering 
\caption{Human Evaluation for  {\sc nosup} (N $=$ 100)
{\sc attr} (N $=$ 100) and {\sc bool} (N $=$ 86) for Naturalness, Semantic Coherence, and Grammaticality }
\label{table:human-evaluation}
\end{table}

\noindent{\bf Human Evaluation.}  The automatic SER results provide 
insight into the semantic accuracy of the models, but no
assessment of other aspects of performance. We thus conduct a human
evaluation on Mechanical Turk to qualitatively assess
fluency, coherency and grammaticality.  We use the automatic SER to
select 100 semantically perfect references from the {\sc
  nosup} and the {\sc attr} models' test outputs, and the 86 perfect
references from {\sc bool}.  We ask 5 Turkers to judge on a scale of
1 (worst) to 5 (best) whether the utterance is: (1) fluent and
natural; (2) semantically coherent; and (3) grammatically well-formed.
Table~\ref{table:human-evaluation} reports the average score for these
qualitative metrics as well as the Turker agreement, using the average Pearson
correlation across the Turkers.  The
results show that the agreement among Turkers is high, and that all the models perform well, but that
the {\sc attr} model outputs are the most natural and coherent, while the {\sc bool} model outputs are the most grammatical.


\subsection{Self-Training}
\label{self-train-sec}

In order to conduct self-training experiments, we need perfect outputs
that combine attributes from both sources to add back into training. These outputs must also be natural, coherent and grammatical, but Table~\ref{table:human-evaluation} shows that this is true of all the models.
A key idea for our novel self-training method is that the
TTM (Section~\ref{ttm-sec}) automatically
produces ``retrofit'' corrected MRs that match the output texts of the
NLG models.  Thus we expect that we can construct more perfect outputs for
self-training by using retrofitting than those in Table 
\ref{table:supervision-evaluation}. Here, we first 
analyse the outputs of the initial models to show that self-training is feasible, and then explain our method and present
results.

 \begin{figure}[h!]
 	\centering
     \includegraphics[width=\columnwidth, keepaspectratio]{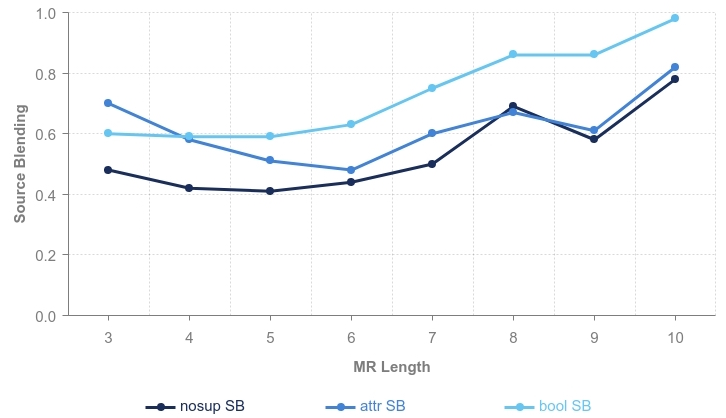}
     \caption{Source Blending Rate (SB) 
 as a function of MR length  for {\sc nosup}, {\sc attr} and {\sc bool}.  \label{fig:sb}}
     \vspace{-.1in}
 \end{figure}

\noindent{\bf Error Analysis.} An initial examination of the 
outputs suggests that the models simply have trouble combining
attributes from both sources. We provide examples
in Table~\ref{table:error-sentences} in
Section~\ref{output-supervision-examples-supplement} in the
supplementary materials.
To quantify this observation, we define a
metric, Source Blending Rate ({\bf SB}), that counts the percentage of
outputs that combine attributes from both sources, whether or not the
attribute values are accurate:
\begin{center}
 ${\displaystyle {\mathit {SB}}={\frac {R_{sb}}{N}}}$
\end{center}
 where $R_{sb}$ is the count of references $r$ that 
 contain an attribute $a_i$  $\subseteq\ source_1$ and another
 attribute $a_j$  $\subseteq\ source_2$, and N is the total number
 of references. Only attributes that appear uniquely in each
 source are included in the $a_i$, $a_j$: the unique attributes
are illustrated
in Figure~\ref{ontology-fig}. 

Figure~\ref{fig:sb} graphs SB as a function of MR length showing that
indeed the models {\bf do} in many cases produce combined outputs and that the type of model supervision greatly influences SB. 
The {\sc nosup} model is the worst: a fact that is masked by the
{\sc nosup} model's SER in Table~\ref{table:supervision-evaluation},
which appears to be on a par with both {\sc attr} and {\sc bool}. 
Interestingly, all models are more likely to produce an SB
output as the MRs get longer, but Figure~\ref{fig:sb} shows clearly that the
{\sc bool} model especially excels.

For self-training, we also need a model that generates utterances with  the {\sc recommend}
dialogue act. 
As mentioned in Section~\ref{ontology-sec}, recommendations increase both semantic and syntactic
complexity. Half  the test items contain a recommendation, so we need a model that can produce them. 
Table~\ref{rec-table} presents results for SER and SB
depending on whether a {\sc recommend} was in the MR, showing that the three models vary a great deal. However, the {\sc
  bool} row for the SB column shows that when the MR includes a
recommendation, the {\sc bool} model produces a combined output far more
frequently than {\sc nosup} or {\sc attr} (SB = .73).


\begin{table}[!htb]
\begin{footnotesize}
\begin{tabular}
{@{} l||cc|cc @{}}
\hline
 \bf Model   &\multicolumn{2}{c|}{\cellcolor[gray]{0.9} {\sc ser}} & \multicolumn{2}{c}{\cellcolor[gray]{0.9} {\sc sb}} \\
& { \cellcolor[gray]{0.9} {\sc rec}} & { \cellcolor[gray]{0.9} {\sc no-rec}} & { \cellcolor[gray]{0.9}  {\sc rec }}  & { \cellcolor[gray]{0.9}  {\sc no-rec }}\\ \hline  
\sc nosup &  .43 & .46  & .44 & .56 \\ 
\sc attr & .51 & .41  & .36  & .77  \\ 
\sc bool & .47 & .43  & .73 &  .67  \\ \hline
\end{tabular}
\end{footnotesize}
\vspace{-.1in}
\centering 
\caption{Effect of the {\sc  recommend} dialogue act on Slot Error Rate (SER) and Source Blending (SB)
for  the three types of model supervision: {\sc nosup}, {\sc attr} and {\sc bool}.
\label{rec-table}}
\end{table}

Thus Figure~\ref{fig:sb} and Table~\ref{rec-table} show that the
{\sc bool} model
produces the most combined outputs. After TTM extraction, the {\sc bool} model
provides the most  instances (1405) of retrofit MR/output pairs to
add to self-training, and we therefore use {\sc bool} in the
self-training experiments below.

\begin{table*}[t!bh]
\begin{footnotesize}
\begin{tabular}{p{2.3in}|p{1.7in}|p{1.8in}} \toprule
 \bf Original MR & \bf Text-to-MR & \bf OUTPUT \\ \hline     
name[{\sc restaurant}], cuisine[fastfood], decor[good], qual[fantastic],~location[riverside], price[cheap], eatType[pub], familyFriendly[no] & name[{\sc restaurant}], cuisine[fastfood], qual[good], location[riverside], familyFriendly[no]	 & [{\sc restaurant}] is a fast food restaurant located in the riverside area. it has good food and it is not family friendly. \\ \hline
name[{\sc restaurant}], recommend[yes], cuisine[fastfood], qual[good], location[riverside], familyFriendly[no] & name[{\sc restaurant}], cuisine[fastfood], qual[good], location[riverside], familyFriendly[no] & [{\sc restaurant}] is a fast food restaurant in the riverside area. it is not family friendly and has good food. \\ 	
\bottomrule
\end{tabular} 
\end{footnotesize}
\vspace{-.1in}
 \caption{\label{self-training-retrofitting-example}  Examples to show retrofitting. The examples
start from different original MRs (col 1), but yield the same MR after text-to-MR extraction (col 2). In Row 1,  the model output in column 3 deleted the attributes {\it price}, {\it decor} and {\it eat type} (pub), and substituted
the value ``good'' for ``fantastic'' for the quality attribute. In  Row 2 the model deleted the {\sc recommend} dialogue act, but otherwise realized the original MR  correctly. At test time, the original MRs produced different outputs (col 3). Thus the retrofitting yields two unique novel instances for self-training.}
\end{table*}

\noindent{\bf Retrofitting MRs for Self-Training.}
Table~\ref{self-training-retrofitting-example} illustrates how the TTM
works, and shows that it can effectively create a new MR that may not
have been previously seen in training,  allowing the model to {\bf learn
from its mistakes}. The caption for
Table~\ref{self-training-retrofitting-example} explains in detail the
retrofitting process and how it leads to new examples to use in
self-training.

It is important to note that the retrofit MRs for some NLG outputs
{\bf cannot} be used for self-training.  NLG model outputs whose
semantic errors include repetitions can {\bf never} be used in
self-training, because valid MRs do not include repeated attributes
and values, and the method doesn't edit the NLG output string.
However, deletion errors cause no issues: the retrofit MR simply
doesn't have that attribute. Substitutions and hallucinations can be
used because the retrofit MR substitutes a value or adds a value to
the MR, as long as the realized attribute value is valid, e.g.
``friendly food'' is not a valid value for {\it food
  quality}.\footnote{We applied the human evaluation
  in Section~\ref{ttm-sec} to instances included in self-training: the correlation between human
  judgements and the automatic SER is .95, indicating that the retrofit MRs are
  highly accurate.}$^{,}$\footnote{Table \ref{table:error-sentences} in
Section~\ref{output-supervision-examples-supplement} provides
additional examples of errorful outputs that {\bf can} or {\bf
  cannot} be used in self-training.}

\noindent{\bf Experiments.} To begin the self-training experiments, we
apply the source-blending metric (SB) defined above to identify
candidates that combine attributes from both sources, and then apply
the TTM to construct MRs that match the NLG model outputs,
as illustrated in Table~\ref{self-training-retrofitting-example},
eliminating references that contain a repetition.
We start with the same combined 76,832 training examples and the 1405
retrofit MR/NLG outputs from the {\sc bool} model.  We explore two
bootstrapping regimes, depending on whether a model output is a
repetition of one that we have already seen in training. One model
keeps repetitions and adds them back into training, which we dub
S-Repeat, and the other model only adds unique outputs back into
training, which we dub S-Unique. 

\begin{figure}[!bht]
	\centering
    \includegraphics[width=\columnwidth, keepaspectratio]{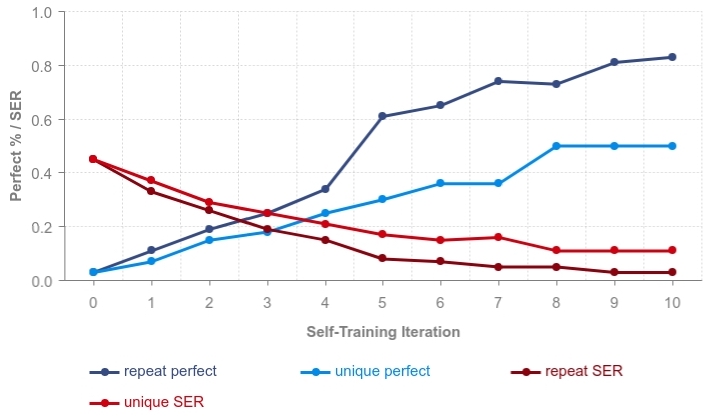}
    \vspace{-.2in}
\hspace{.5in}\caption{SER and perfect\% on the COM test set for S-Repeat vs. S-Unique during self-training}
    \label{fig:perfect-self-train}
\end{figure}

\noindent{\bf Quantitative Results.} 
Figure \ref{fig:perfect-self-train} shows how the SER and perfect\%
continuously improve on the COM test set
for S-Repeat over 10 rounds of self-training, and
that S-Repeat has better performance, indicating that adding
multiple instances of the same item to training is useful. The
performance on the COM test set of the S-Unique model flattens after 8 rounds. After 10
rounds, the S-Repeat model has an SER of .03 and produces perfect
outputs 82.9\% of the time, a 77.4 percent absolute improvement over the best results in Table~\ref{table:supervision-evaluation}. 

\noindent{\bf COM-2 Test Set.} Since the  the self-training procedure used the COM test set during self-training, we construct a new test with 3040 novel MRs using the procedure  
described in Section~\ref{ontology-sec}, which we  call COM-2. First we test  the initial models on COM-2, resulting in a best SER of 0.45 for the {\sc bool} model, identical with the result for  COM. For perfect\% the best result was 5.3\% on 
the {\sc attr} model, which is again comparable to the original COM test set. We then tested the final self-trained model  on COM-2, with the result that  the SER for  S-Repeat (0.03) and S-Unique (0.11) are  again identical to the result for COM.  The perfect\% is comparable to 
that reported in Figure \ref{fig:perfect-self-train};
it decreases by 2.2\% 
for S-Repeat to 80.7\% and increases by .2\% for S-Unique to 50.7\%. Overall, 
the performance on COM-2 improved by an absolute 75.4\%.

\begin{figure}[!bht]
	\centering
    \includegraphics[width=\columnwidth,keepaspectratio]{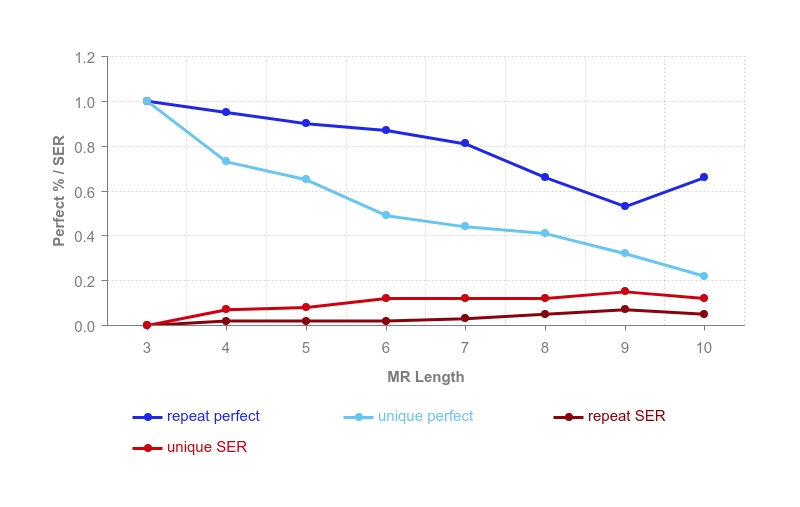}
    
    \vspace{-.2in}
\hspace{.5in}\caption{SER and perfect\% on COM-2 as a function of MR length  for {\sc bool} supervision before self-training and for the S-Repeat model after self-training.}
    \label{fig:ser-perf-self-training}
\end{figure}

Figure~\ref{fig:ser-perf-self-training} shows that the results
improve, not only overall, but also by MR length.
It plots the SER and perfect\%
results, by MR length, for the {\sc bool} model before and after
self-training. 
While the perfect\% decreases as the number of
attributes increase, there is a large improvement over the initial 
model results. Also, after
self-training the worst perfect\% is still above 0.5, which is higher
than perfect\% for any MR length before self-training. The SER also
improves over all MR lengths after self-training, not exceeding .06,
significantly better than even the shortest MR before self-training.\footnote{Performance results for the self-trained model on the original E2E and NYC test sets in supplement A.3 shows that 
performance also improves on the E2E and NYC test sets.}



\noindent{\bf Human Evaluation.}  
\begin{table}[!htb]
\begin{footnotesize}
\begin{tabular}
{@{} l||ccc @{}}
\hline
 \bf Model   &{ \cellcolor[gray]{0.9} {\sc Nat.}} & { \cellcolor[gray]{0.9} {\sc Coher.}} & { \cellcolor[gray]{0.9}  {\sc Grammat. } }
\\ \hline  
\sc S-Repeat & 3.99 &  4.08 & 4.02 \\ 
\sc S-Unique & \bf 4.06  &  \bf 4.13  &   \bf 4.14  \\ 
\sc agreement & .57 & .61 & .57 \\ 
\hline
\end{tabular}
\end{footnotesize}
\vspace{-.1in}
\centering 
\caption{Human Evaluation on Mechanical Turk for  S-Repeat (N $=$ 100) and S-Unique (N $=$ 100) for Naturalness, Semantic Coherence, and Grammaticality }
\label{table:strain-human-evaluation}
\end{table}
We also performed a human evaluation on Mechanical Turk to assess the qualitative properties of the model outputs after self-training.  We selected 100 perfect references for S-Repeat and 100 for S-Unique and used the same HIT as described in Section~\ref{model-results-sec}.
Table~\ref{table:strain-human-evaluation} reports the average score for these qualitative metrics as well as the Turker agreement, using the average Pearson correlation across the Turkers. The results show that naturalness, coherence
and grammaticality are still high after self-training for both models,
but that the S-Unique model produce better outputs
from a qualitative perspective. We believe we could improve the self-training method used here with additional referenceless evaluation metrics that aim to measure naturalness and grammaticality \cite{MehriEskenazi-FED}. We leave  this to future work. 

\begin{table}[h!tb!]
\begin{footnotesize}
\begin{tabular}
{@{} p{0.2cm}|p{6.3cm} @{}} \toprule
\bf \#  & \textbf{Realization}  \\ \midrule 
1 & [{\sc restaurant}] is {\color{darkred}the best place} because it is a {\color{blue}family friendly pub} with {\color{darkred}good decor} and {\color{darkred}good food}. \\ \hline
 2 & [{\sc restaurant}] is a {\color{blue}family friendly} restaurant with {\color{darkred}bland food} and is in the low price range. It is {\color{darkred}the best restaurant}.\\ \hline
3 & [{\sc restaurant}] is a {\color{blue}family friendly coffee shop} with {\color{darkred}decent service} and a {\color{blue}low customer rating}. It is in the \pounds20-25 price range. \\ \hline
4 & [{\sc restaurant}] is {\color{darkred}the best restaurant} because it is in the east village, it is {\color{blue}near [{\sc point-of-interest}]} with {\color{darkred}great service} and it is affordable.\\ \hline
\end{tabular}
\end{footnotesize}
\vspace{-.05in}
\caption{Example outputs with source blending. NYC attributes are represented  using {\color{darkred}red} and E2E attributes are represented using {\color{blue}blue} \label{table:qual-example-sentences}}.
\end{table}

\noindent{\bf Qualitative and Linguistic Analysis.} Table~\ref{table:qual-example-sentences} provides outputs from the models that display different ways of combining
attributes from the original sources. 
In Row 1 we can see that the {\sc recommend} dialogue act from NYC can be 
combined in the same sentence as the attributes {\it family friendly} and 
{\it eat type} from E2E and aggregate
these E2E attributes with NYC attributes {\it decor} and {\it food quality} using a ``with" operator. Row 2 shows 
another example where the NYC and E2E attributes are joined using a 
``with" operator. In Row 3 there is a single sentence with 
four attributes where the NYC attribute is preceded and 
followed by E2E attributes. 
Row 4 
concatenates the two sources in a single sentence using sentence coordination. The ``east village" location from  the NYC dataset, is 
concatenated with the attributes {\it near} from E2E and {\it service} from NYC.   These examples 
show that the NLG models can combine attributes from both sources in many  different ways. Table~\ref{table:st-example-sentences} in Section~\ref{output-final-examples-supplement} provides additional detail by providing examples along with their corresponding MRs.


\section{Conclusion}
\label{conc-sec}

\nocite{KedzieMcKeown19,shah2018bootstrapping}
\nocite{budzianowski2018multiwoz,eric2019multiwoz,gavsic2015policy,hakkani2016multi,Cervoneetal19,shah2018bootstrapping,ultes2017pydial,chen2017deep}

This paper presents the first experiments on training an NLG for an
extended domain ontology by re-using existing within-domain training
data.  We show that we can combine two training datasets for the
restaurant domain, that have different ontologies,
and generate output that
combines attributes from both sources, by applying a combination of
neural supervision and a novel self-training method.  While it
is common practice to construct test sets with unseen attribute
combinations, we know of no prior work based on
constructing a new combined ontology. Our experiments show that the
task is surprisingly adversarial, consistent with recent work
suggesting that neural models often fail to generalize
\cite{Wallaceetal19,Fengetal18,Ribeiroetal18,goodfellow2014explaining}. 
Work on  domain transfer shares similar goals to the experiments presented here 
\cite{wen2016multi,golovanov2019large}, but these methods
do not produce NLG outputs that integrate attributes from
two different sources into the same sentence. Our final results show
that the ability of our self-training method to automatically construct new training
instances  results in high quality natural, coherent and
grammatical outputs with high semantic accuracy.
  
In future, we hope to generalize our novel self-training method
to build an NLG that can combine two distinct domains, e.g. 
hotels or movies combined with restaurants in multi-domain dialogue
\cite{budzianowski2018multiwoz,gavsic2015policy,hakkani2016multi,Cervoneetal19,ultes2017pydial}. Ideally
systems that cover multiple domains should be able to produce
utterances that seamlessly integrate both domains, if data exists for
each domain independently.  However, there may be additional 
challenges in such combinations. Our results require the initial neural models to generate {\bf some} combined
outputs. It is not clear whether there are some aspects of our experimental setup that facilitate this, e.g. it may require  some attributes to be
shared across the two initial ontologies, or  some shared
vocabulary. Thus it is possible that initial models for two more
distinct domains may not produce any combined outputs, and it may
be necessary to seed the self-training experiments with a small number of
combined training instances. We leave these issues to future work.





\bibliography{nl}
\bibliographystyle{acl_natbib}


\clearpage
\appendix
\setcounter{page}{1}

\section{Supplementary Materials: Learning from 
Mistakes: Combining Ontologies via Self-Training for Dialogue Generation}

\subsection{Types of Semantic Errors}
\label{semantic-error-supplement}


The TTM is tuned to identify 4 common neural generation
errors: {\it deletions} (failing to realize a value), {\it repetitions}
(repeating an attribute), {\it substitutions} (mentioning an attribute
with an incorrect value), and {\it hallucinations} (introducing an
attribute that was not in the original MR at all).

Table~\ref{table:semantics-sentences} illustrates each of these types of semantic
errors. Row 1 shows deletions of {\it cuisine}, {\it price} and
{\it near} which are in the MR but not in the 
realization. Row 2 demonstrates a repetition, 
where {\it location} and {\it decor} are both repeated. 
{\it Decor} is realized with two different lexical values, 
``good ambiance" and ``good decor".
There is a substitution in Row 3 where the MR
states that the {\it food quality} is ``bad'', but {\it food quality} is realized as 
''good''. Finally, Row 4
has a hallucination, 
{\it service} is not in the MR but it in the 
second sentence of the realization.

\subsection{Example Errorful NLG Model Outputs} 
\label{output-supervision-examples-supplement}

Table \ref{table:error-sentences} provides 
examples of NLG model output utterances with high SERs. It illustrates how the NLG models struggle to combine attributes from the two ontologies which
is required by all the input MRs (Column SB). It also illustrates cases where it is not possible to produce a valid retrofit MR that can be added back into training during self-training (Column Valid). In most cases these
are due to many repetitions. 
Row 1 is an example where there is no 
source blending and
since it has a repetition ({\it price}) it cannot be
used for self-training (valid = no). Row 1 also illustrates an ungrammatical realization of {\it price} which we have no way to automatically detect at present {\it it is in the high price.}  Row 2 has three 
deletions as well as two repetitions. The output repeats {\it It is in midtown}
three times in a row. Row 3 has five 
errors, it does not realize the dialogue act {\sc recommend}
and has deleted three other attributes and it hallucinations 
{\it food quality}. While this is a significant number of errors, 
this realization can still be used in self-training, since 
none of its errors are repetitions. Row 4 has all four types 
of errors. It deletes {\it cuisine}, {\it
decor} and {\it service}, it realizes a value for  
{\it family friendly} twice with different values, a substitution and finally  
it hallucinates {\it food quality}. Row 5
actually has more errors than slots. It deletes all but two of 
its attributes: {\it name} and {\it rating}. It also hallucinates 
{\it food quality} and repeats {\it rating}.

\begin{table}[!ht]
\begin{footnotesize}
\begin{tabular}
{@{} l|l|l||r|r @{}}
\hline
\bf Model & \bf Training  & \bf Test   &{ \cellcolor[gray]{0.9} {\sc ser}} & { \cellcolor[gray]{0.9} {\sc perf} \%}
\\ \hline  
\sc bool & \sc S-Repeat & \sc e2e & .14 & 25\% \\
\sc bool & \sc S-Repeat & \sc nyc & .05 & 77\% \\
\hline
\end{tabular}
\end{footnotesize}
\vspace{-.1in}
\centering 
\caption{
Performance of the self-trained S-Repeat model on the original E2E and NYC test sets.  
E2E test N = 630. NYC test N = 314.}
\label{table:self-baselines-combo-nocombo}
\end{table}

\subsection{Performance on E2E and NYC test sets}

Table~\ref{table:baselines-combo-nocombo} provided a 
baseline for {\sc nosup}'s
performance before self-training on 
the original test sets for E2E and NYC. We also verify that the self-trained model performs well after self-training. 
Table~\ref{table:self-baselines-combo-nocombo}
shows that self-training improves the results for the original E2E and NYC test sets. 


\begin{table*}[!b]
\begin{footnotesize}
\begin{tabular}
{@{} p{2.1cm}|p{6.85cm}|p{6.35cm} @{}} \toprule
\textbf{Error Type} & \bf MR & \textbf{Realization}  \\ \midrule 
Delete {\sc cuisine, price, near}  & name[{\sc restaurant}], cuisine[mexican], location[midtown], price[expensive], eatType[coffee shop], familyFriendly[no], near[{\sc point-of-interest}] & [{\sc restaurant}] is a coffee shop that is not family friendly. It is located in Midtown.  \\ \hline
Repeat {\sc location}, {\sc decor}  & name[{\sc restaurant}], decor[good], location[midtown west], eatType[coffee shop], rating[1 out of 5] & [{\sc restaurant}] is a coffee shop in Midtown West with good ambiance. It is in Midtown West with good decor. \\ \hline
Substitution {\sc quality bad} to {\sc good}  & name[{\sc restaurant}], decor[good], qual[bad], location[tribeca/soho], eatType[pub] & [{\sc restaurant}] is in Tribeca/Soho with good food and good decor. It is a pub.\\ \hline
Hallucination {\sc service}  & name[{\sc restaurant}], decor[good], qual[good], location[riverside], near[{\sc point-of-interest}] & [{\sc restaurant}] is near [{\sc point-of-interest}] in the riverside area. It has good food, good decor and good service. \\ \hline
\end{tabular}
\vspace{-.1in}
\caption{Model outputs illustrating types of semantic errors.
\label{table:semantics-sentences}}
\end{footnotesize}
\end{table*}

\begin{table*}
\begin{footnotesize}
\begin{tabular}
{@{} p{0.1cm}|p{0.4cm}|p{0.5cm}|p{0.3cm}| p{6.3cm}|p{6.3cm} @{}} \toprule
\bf \# & SER & Valid & SB &  \bf MR & \textbf{Realization}  \\ \midrule 

1 & .60 & no & no &
name[{\sc restaurant}], {\tt decor[fantastic]}, {\tt qual[acceptable]}, price[high], {\it familyFriendly[yes]} & [{\sc restaurant}] is a child friendly restaurant with french food and it is in the high price range. it is in the high price. \\ \hline	

 2 & .63 & no& no
& name[{\sc restaurant}], {\tt recommend[yes]}, cuisine[japanese], {\tt decor[bad]}, location[midtown], {\tt service[fantastic]}, {\it rating[low]}, {\it near[{\sc point-of-interest}]} & 
[{\sc restaurant}] is the best restaurant since it is a japanese restaurant with bad ambiance and it is in midtown. it is in midtown. it is in midtown.\\ \hline

 3 &  .71 & yes & yes
& name[{\sc restaurant}], {\tt recommend[yes]}, {\tt decor[fantastic]}, {\tt service[acceptable]}, {\it eatType[pub]}, {\it rating[high]}, {\it near[xnear]} &
[{\sc restaurant}] is a pub with good food and it is located near the xnear.

\\ \hline
  4 & .75 & no & yes &  name[{\sc restaurant}], cuisine[southern], {\tt decor[fantastic]}, location[city centre], price[cheap], {\tt service[good]}, {\it eatType[coffee shop]}, {\it familyFriendly[no]}
& [{\sc restaurant}] is a cheap, family friendly coffee shop with good food. it is in the city centre. it is not family friendly. 
\\ \hline
5 & 1.17 & no & yes
& name[{\sc restaurant}], cuisine[indian], {\tt decor[fantastic]}, {\tt service[good]}, {\it rating[high]}, {\it near[{\sc point-of-interest}]} & 
[{\sc restaurant}] provides good food and has great customer rating and has great customer rating .

\\ \hline

\end{tabular}
\vspace{-.25in}
\caption{Example pathological outputs with high SERs from the NLG models before self-training. Valid realizations (col 3) are defined as those usable for self-training. In the MRs, the NYC attributes are represented using {\tt typewriter font} and E2E attributes are represented using {\it italics}. \label{table:error-sentences}}.

\end{footnotesize}
\end{table*}

\begin{table*}[!htb]
\begin{footnotesize}
\begin{tabular}
{@{} p{0.2cm}|p{7.7cm}|p{7.3cm} @{}} \toprule
\bf \# & \bf MR & \textbf{Realization}  \\ \midrule 
 1 & name[{\sc restaurant}],cuisine[Indian], {\tt decor[fantastic]},{\tt qual[good]},  {\tt service[good]}, {\it rating[high]}, {\it near[{\sc point-of-interest}]} &  
  [{\sc restaurant}] is a Indian restaurant with great service and excellent ambiance and a high customer rating. It is located near the  [{\sc point-of-interest}]. \\ \hline
 2 & name[{\sc restaurant}],  {\tt recommend[yes]}, {\tt decor[good]}, {\tt qual[good]}, price[cheap], {\tt service[bad]}, {\it familyFriendly[yes]},& [{\sc restaurant}] is a family friendly restaurant with good food, good ambiance and bad service. It is in the low price range.  \\ \hline
  3 & name[{\sc restaurant}], {\tt recommend[yes]}, {\tt decor[bad]}, {\tt qual[good]}, location[flatiron/union square], price[cheap], {\tt service[acceptable]}, {\it eatType[coffee shop]}, {\it rating[3 out of 5]}, {\it } & [{\sc restaurant}] is the best restaurant because it is a family friendly coffee shop with good food, friendly service and bad ambiance. It is in Flatiron/Union Square. It has a customer rating of 3 out of 5. It is cheap.\\ \hline
4 & name[{\sc restaurant}], {\tt recommend[yes]},
cuisine[mediterranean], {\tt decor[fantastic]}, price[very expensive],
{\it eatType[pub]}, {\it rating[5 out of 5]} & [{\sc restaurant}] is a
Mediterranean pub with excellent ambiance and a customer rating of 5
out of 5. It is in the upper price range. It is the best restaurant.\\ \hline

\end{tabular}
\vspace{-.25in}
\caption{Example outputs of source blending from final self-training iterations. In the MRs, the NYC attributes are represented using {\tt typewriter font} and E2E attributes are represented using {\it italics}. \label{table:st-example-sentences}}.
\end{footnotesize}
\end{table*}


\subsection{Example Final Model Outputs} 
\label{output-final-examples-supplement}

Table~\ref{table:st-example-sentences} provides outputs from the final
iteration of self-training that display different ways of combining different
attributes from the ontologies. Row 1 shows that the model can combine attributes from the
two sources in the same sentence, with attributes from each source,
{\it decor} and {\it rating}, appearing in a single sentence with {\it
  and}.
  Row 2 shows a different way of combining attributes from the
two sources, with {\it family friendly} and {\it food quality}, in a
single sentence, this time using {\it with}.  In Row 3 we can see that
the model can also generate complex sentences for recommendations  using the marker {\it because}.  Also, the attribute used in the {\it because} clause  is from
E2E i.e. {\it family friendly} but such sentences never appear in the
original E2E training data. The last row shows a complex sentence
where {\it decor} is combined with {\it eat type} and {\it customer rating},
again a novel combination.

\end{document}